\documentclass[a4paper]{article}
\usepackage{balance}   

\usepackage{INTERSPEECH2022}
\usepackage[ruled,vlined,linesnumbered]{algorithm2e}
\DeclareMathOperator*{\minimize}{minimize}

\title{Unsupervised Speaker Diarization that is Agnostic to Language, Overlap-Aware, and Tuning Free}
\name{M. Iftekhar Tanveer$^1$, Diego Casabuena$^1$, Jussi Karlgren$^2$, Rosie Jones$^1$}
\address{
  $^1$Spotify, USA \& 
  $^2$Spotify, Sweden
  }
\email{\{iftekhart, diegoc, jkarlgren, rjones\}@spotify.com}

\begin{document}

\maketitle
\begin{abstract}
Podcasts are conversational in nature and speaker changes are frequent---requiring speaker diarization for content understanding. We propose an unsupervised technique for speaker diarization without relying on language-specific components. The algorithm is overlap-aware and does not require information about the number of speakers. Our approach shows $79\%$ improvement on purity scores ($34\%$ on F-score) against the Google Cloud Platform solution on podcast data.
\end{abstract}

\section{Introduction}
\emph{Speaker Diarization} is the process of logging the timestamps (or writing in a diary---thus the term ``diarization'') when different speakers spoke in an audio recording. At Spotify, we have found that reliable speaker diarization is a foundational technology we need in order to build downstream natural language understanding pipelines for podcasts. Podcasts are more diverse in content, format, and production styles~\cite{jones2020trec, jonesEA2021currentchallenges, clifton2020hundredthousand, karlgren2021trec} than classical application areas for diarization---requiring a revisit to the techniques.

\begin{figure}[t]
  \centering
  \includegraphics[width=0.9\linewidth]{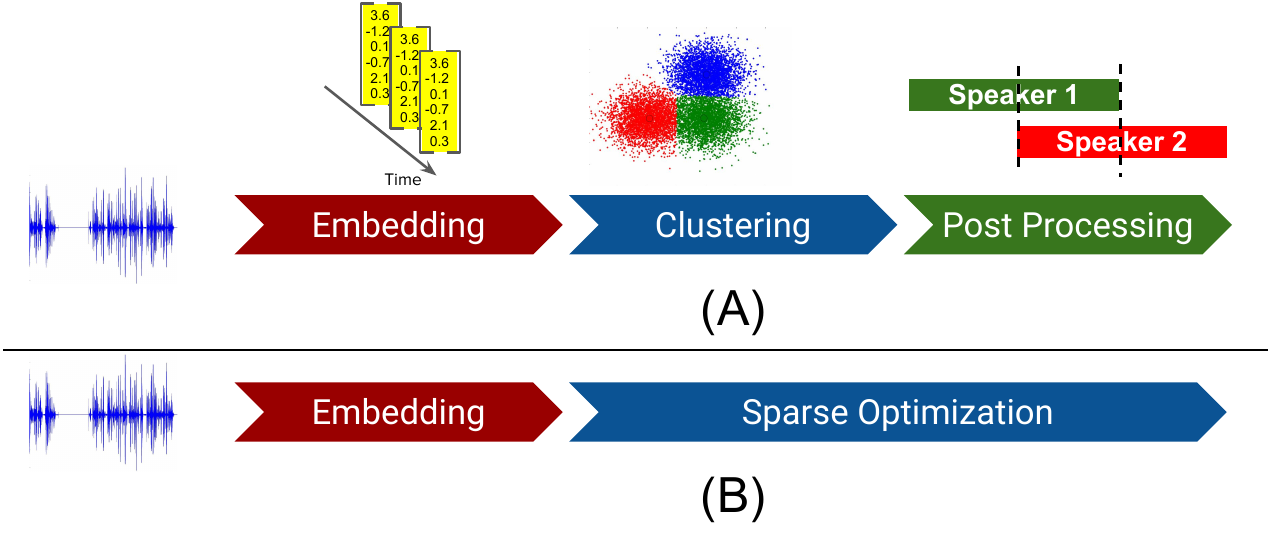}
  \caption{(A) Classical speaker diarization approach vs. (B) our proposed approach}
  \label{fig:diarization_pipeline}
\end{figure}

Many established diarization techniques utilize a pipeline similar to the one shown in Figure~\ref{fig:diarization_pipeline}(A) ~\cite{park2022review}. A typical pipeline involves several independent components. The \emph{Embedding} component segments the audio into small chunks and converts each chunk into a vector (e.g. i-vectors~\cite{dehak2010front}, d-vectors~\cite{heigold2016end}, or x-vectors~\cite{snyder2018x}) which represents the auditory characteristics of the speech heard in a chunk and the speakers who produce it. The \emph{Clustering} component divides those vectors into several groups by applying Spectral Clustering~\cite{ning2006spectral, shum2013unsupervised, wang2018speaker}, Agglomerative Clustering~\cite{sell2018diarization}, or other techniques~\cite{kenny2010diarization, valente2010variational}. Finally, a \emph{Post Processing} component takes care of some issues not handled by the previous components, such as detecting overlapping speech with more than one simultaneous speaker~\cite{bullock2020overlap}. 

Classical research in speaker diarization was focused on independently improving the performances of these blocks~\cite{park2022review}. However, with growing popularity in the end-to-end~\cite{fujita2019end} learning models using e.g. deep learning, recent implementations have used combinations of multiple modalities (e.g. audio and text)~\cite{dey2018end}. For instance, Google has developed an algorithm that performs speech recognition jointly with diarization~\cite{shafey2019joint}; d-vectors have been proposed to utilize both audio and text~\cite{heigold2016end}.

A drawback of incorporating textual information into a diarization algorithm is it also renders the algorithm dependent on language. Such an algorithm needs to be retrained for every supported language which quickly becomes impracticable\footnote{As of March 8\textsuperscript{th}, 2022, e.g., Google cloud transcription supports speaker diarization for only a handful languages among the languages supported: https://cloud.google.com/speech-to-text/docs/languages}. Human perception is quite capable to perform speaker diarization without understanding the language, and  diarization would seem to be a task which should be solvable automatically without specific training or specific knowledge resources for the language being spoken. In this paper, we propose an audio-only solution for speaker diarization which is naturally \emph{language agnostic}.

The clustering component introduces several problems for the overall diarization effort. Typical clustering algorithms depend on a pre-determined number of clusters---in this application scenario, number of distinct speakers. However, it is impractical to decide the number of speakers beforehand in the speaker diarization scenario. Additionally, many clustering algorithms group the audio segments into 
disjoint clusters. This approach makes identifying overlapping speech (a frequent occurrence in natural and unscripted conversation, such as in podcasts) an afterthought, typically solved by introducing post-processing modules based on e.g. Hidden Markov Models~\cite{boakye2008overlapped}, neural networks~\cite{geiger2013detecting}, or other methods~\cite{raj2021dover}. Fully supervised speaker diarization approaches have been proposed~\cite{zhang2019fully}, but supervision requires manual annotation of large training data sets---which is costly, time-consuming, and dreary work for human annotators. In addition, a training dataset must be balanced over various languages, production styles, and format categories (e.g. audiobooks, enacted drama, panel discussions etc.) which adds up to being a considerable engineering effort.

We propose an unsupervised speaker diarization algorithm that avoids all the problems described above by replacing the clustering and the post-processing components with a sparse ($\ell_1$ regularized) optimization approach as shown in Figure~\ref{fig:diarization_pipeline}(B). We use a pretrained audio embedding model with convenient characteristics that makes the algorithm \emph{overlap aware}. The hyperparameters required for our proposed processing model and algorithm are automatically adjusted by employing theories from compressed sensing~\cite{candes2008restricted, candes2006stable} and linear algebra. Our algorithm is similar to Vipperia et al.~\cite{vipperla2012speech} but unlike that, the proposed algorithm doesn't need any manual intervention for constructing a \emph{global} basis for overlap handling. The embedding basis is constructed automatically by utilizing an $\ell_1$-regularization over a \emph{local} basis matrix. This approach makes it completely \emph{tuning-free} from the users' perspective---enabling the usage in large scale setting as in Spotify. The proposed solution is also highly scalable with hardware such as Graphical Processing Units (GPUs). 

Finally, we compare the performance of the proposed algorithm against several industry standard solutions such as Google cloud platform and PyAnnote. 







\section{Methods}
Our proposed approach has two major steps: a) Construction of the embedding signal, and b) Sparse optimization. 

\subsection{Constructing the Embedding Signal}\label{sec:constructingembedding}
We generate a sequence of $M$-dimensional ($M = 512$) vectors for an audio recording using a pre-trained VggVox~\cite{nagrani2020voxceleb} embedder. VggVox is a convenient choice for our purposes, for several reasons. It is trained on the VoxCeleb~\cite{nagrani2020voxceleb} dataset which contains over 1 million utterances for 6,112 celebrities from around the world in various contexts and situations. Due to the ``wild'' nature of the VoxCeleb dataset, VggVox is able to capture not only the acoustic characteristics, but also dialectal and stylistic speech characteristics. This allows us to use vector similarity metrics (such as cosine similarity) to identify whether two speech segments are from the same speaker.

VggVox vectors have the attractive and convenient characterstic of adhering to a \emph{linearity constraint}---i.e. a VggVox embedding, $\mathbb{V()}$, of an audio chunk from speaker $S_1$ of length $p$ concatenated with an audio chunk from speaker $S_2$ of length $q$, $\mathbb{V}(S_1[1:p]\oplus S_2[1:q])$ is approximately equal to its weighted arithmetic average:

\begin{equation}
    \frac{1}{p+q}(p\mathbb{V}(S_1[1:p])+q\mathbb{V}(S_2[1:q]))
    \label{eq:vggvox_linearity}
\end{equation}

We segment a recording into a series of overlapping chunks by sliding a $6$-second window over the recording with a variable step size of $1$ second or less. We set the step size to be such that it yields at least 3,600 chunks and compute a vector for each using the pretrained VggVox model. 


We use MobileNet\footnote{A.k.a. YAMNet: https://tfhub.dev/google/yamnet/1}\cite{howard2017mobilenets} to detect non-speech regions in the recording and we set the vectors for non-speech regions to zero. We arrange the sequence of these vectors for a recording into a matrix which we call the \emph{embedding signal}, $\mathbf{\mathcal{E}}\in \mathbb{R}^{M \times T}$ so that vectors from each of the $T$ time-steps are represented in the columns of the matrix. We normalize the embedding signal, $\mathbf{\mathcal{E}}$, such that the columns are unit vectors.

\subsection{Speaker Diarization using Sparse Optimization}
\begin{figure}[t]
  \centering
  \includegraphics[width=0.95\linewidth]{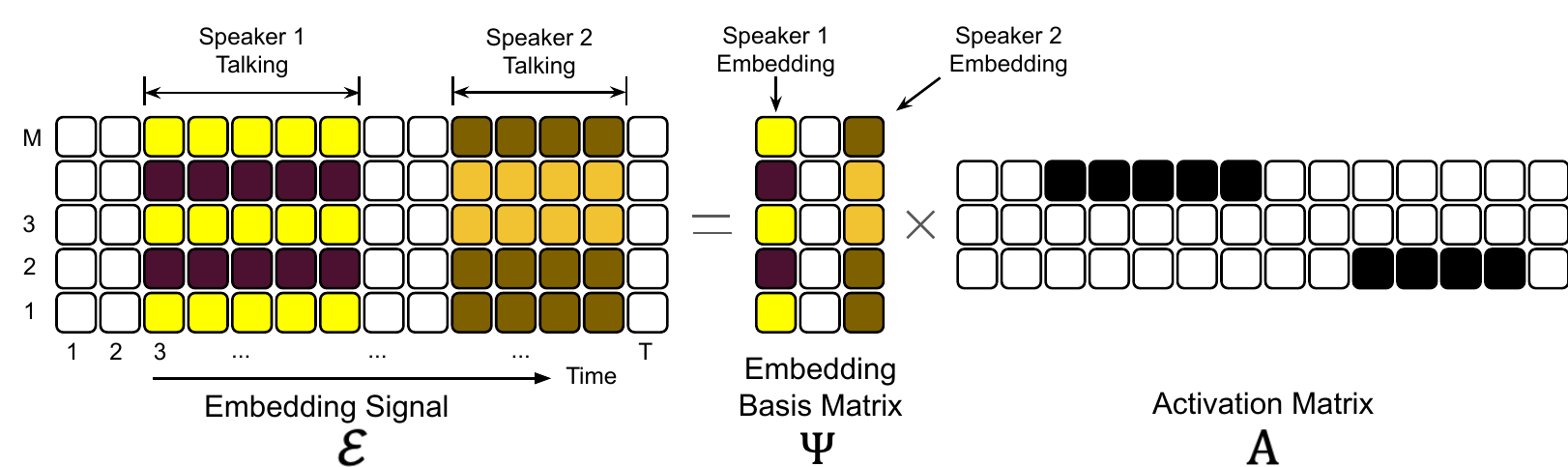}
  \caption{Model of the Embedding Signal as a matrix product of Embedding Basis Matrix and Activation Matrix}
  \label{fig:embedding_signal_model}
\end{figure}

The embedding signal, $\mathbf{\mathcal{E}}$, has a useful property that allows us to use techniques from compressed sensing~\cite{candes2008restricted, candes2006stable} for speaker diarization. The embedding signal is, in general, low rank with many dependent columns. This is due to the fact that the step size of the sequence is short and speakers do not typically change turns at a rate higher than the step size which means that the columns of $\mathbf{\mathcal{E}}$ remain identical over several consecutive time-steps. It is possible to factor out $\mathbf{\mathcal{E}}$ as a matrix product of an \emph{embedding basis matrix}, $\mathbf{\Psi}$ and an \emph{activation matrix}, $\mathbf{A}$ as shown in Figure~\ref{fig:embedding_signal_model}; or, mathematically as in equation~\ref{eq:embedding_signal_model}.
\begin{equation}
\mathbf{\mathcal{E}} = \mathbf{\Psi}\mathbf{A}
\label{eq:embedding_signal_model}
\end{equation}
where the columns of $\mathbf{\Psi} \in \mathbb{R}^{M\times k}$ represent the $M$-dimensional embeddings for each speaker in the audio; $k$ is the maximum number of allowed speakers, and $T$ is the length of the embedding signal, i.e. the number of timesteps. The rows of $\mathbf{A} \in \mathbb{R}^{k\times T}$ represent which speaker is activated at a certain timestep. The elements of $\mathbf{A}$ can take a value in $0$ to $1$. This model converts the diarization problem into a matrix factorization problem. This formulation does not require us to know the \emph{exact number} of the speakers in the audio. As long as the value of $k$ is large enough, this formulation works by setting the activation and embedding for all the unused speakers to zero.

\subsubsection{Formulating the optimization problem}
We solve the matrix factorization problem using a sparse optimization approach. We minimize the $\ell_1$-norm\footnote{The $\ell_1$-norm of $\mathbf{X} \in \mathbb{R}^{U,V}$, $\|\mathbf{X}\|_1:= \sum_{u=1}^U\sum_{v=1}^V |\mathbf{X}[u,v]|$} of the difference, $\|\mathbf{\mathcal{E}} - \mathbf{\Psi}\mathbf{A}\|_1$, to enforce model constraint as in equation~\ref{eq:embedding_signal_model}. In order to obtain the unique solution of this under-determined problem, we enforce sparsity constraint over both $\mathbf{\Psi}$ and $\mathbf{A}$ by minimizing their $\ell_1$ norms, respectively. From the diarization perspective, the sparsity constraint over $\mathbf{\Psi}$ enforces that the embedding signal be reconstructed by utilizing as few speakers as possible. On the other hand, sparsity over $\mathbf{A}$ enforces the solution contains as few nonzero elements for $\mathbf{A}$ as possible. Furthermore, we also enforce that columns of $\mathbf{\Psi}$ are each within a unit disk and the elements of $\mathbf{A}$ is within the range $[0,1]$.

The exact values of the embeddings may vary slightly from time to time depending on the background music or noise in the audio. To address this concern, we introduce a \emph{jitter loss} that enforces continuity and penalizes producing too many broken values in the rows of $\mathbf{A}$. This loss is expressed as an $\ell_1$-norm of the difference between consecutive values in the rows of $\mathbf{A}$. The overall optimization problem is shown below:
\begin{align}
\minimize_{\text{w.r.t}\;\mathbf{\Psi},\; \mathbf{A}}\quad &\|\mathbf{\mathcal{E}} - \mathbf{\Psi}\mathbf{A} \|_1 +\lambda_1\|\mathbf{\Psi}\|_1+\lambda_2\|\mathbf{A}\|_1 +\lambda_3 \mathbf{J}
\label{eq:optimization_objective}
\end{align}
\begin{align*}
\text{subject to,}\quad & \forall_{r, t}\; 0.0 \le \mathbf{A}[r,t] \le 1.0\\
\text{and,}\quad & \forall_{t} \|\mathbf{\Psi}[:,t]\|_2 \le 1.0\\
\text{where, jitter loss,}\quad & \mathbf{J} = \frac{1}{kT}\sum_{r = 1}^k \sum_{t=2}^T |\mathbf{A}[r, t] - \mathbf{A}[r, t-1]| 
\end{align*}

\subsubsection{Solving the optimization problem}
The objective function shown in equation~\ref{eq:optimization_objective} is non-convex. However, when either $\mathbf{\Psi}$ or $\mathbf{A}$ is held fixed, it becomes convex over the other parameter. We solve this optimization problem by alternatingly updating the two model parameters. We utilize the \emph{fast iterative shrinkage thresholding algorithm}~\cite{beck2009fast} (FISTA) to enforce sparsity over the model parameters, $\mathbf{\Psi}$ and $\mathbf{A}$. In addition, we project the parameters in their respective feasibility space at every iteration. We implement the optimization algorithm in Tensorflow\textregistered~to take advantage of the automatic gradient computation using the Wengert~\cite{wengert1964simple} list\footnote{A.k.a. \emph{Gradient Tape} in Tensorflow} approach. As a result, we do not need to analytically compute the gradients of the objective function. Similar functionality is also available in PyTorch\textregistered. We use Adam~\cite{kingma2014adam} for updating the model parameters. The overall optimization process is shown in Algorithm~\ref{algo:diarization}.
\begin{algorithm}[t]
 \KwIn{$\mathbf{\mathcal{E}}$, $k$}
 \KwOut{$\mathbf{\Psi}$, $\mathbf{A}$}
  $i \leftarrow 0$\;
  $\mathbf{A}\leftarrow$ random, $\mathbf{\Psi} \leftarrow $ random\;
 \While{not converge}{
    Compute loss, $\mathbf{L} = \|\mathbf{\mathcal{E}} - \mathbf{\Psi}\mathbf{A} \|_1 +\lambda_1\|\mathbf{\Psi}\|_1+\lambda_2\|\mathbf{A}\|_1 +\lambda_3 \mathbf{J}$\;
    Calculate gradient of $\mathbf{L}$ w.r.t $\mathbf{\Psi}$, $\mathop{\nabla}_{\mathbf{\Psi}}\mathbf{L}$\;
    Update $\mathbf{\Psi}$ using Adam: $\mathbf{\Psi}^{(i+1)}\leftarrow \mathbf{\Psi}^{(i)}-\gamma_\Psi \mathop{\nabla}_{\mathbf{\Psi}}\mathbf{L}$ \;
    $\mathbf{\Psi}^{(i+1)}\leftarrow {\bf shrink}(\mathbf{\Psi}^{(i+1)})$\;
    $\mathbf{\Psi}^{(i+1)}\leftarrow {\bf project_{unit disk}}(\mathbf{\Psi}^{(i+1)})$\;
    Recompute loss, $\mathbf{L} = \|\mathbf{\mathcal{E}} - \mathbf{\Psi}\mathbf{A} \|_1 +\lambda_1\|\mathbf{\Psi}\|_1+\lambda_2\|\mathbf{A}\|_1 +\lambda_3 \mathbf{J}$\;
    Calculate gradient of $\mathbf{L}$ w.r.t $\mathbf{A}$, $\mathop{\nabla}_{\mathbf{A}}\mathbf{L}$\;
    Update $\mathbf{A}$ using Adam $\mathbf{A}^{(i+1)}\leftarrow \mathbf{A}^{(i)}-\gamma_A \mathop{\nabla}_{\mathbf{A}}\mathbf{L}$ \;
    $\mathbf{A}^{(i+1)}\leftarrow {\bf shrink}(\mathbf{A}^{(i+1)})$ \;
    $\mathbf{A}^{(i+1)}\leftarrow {\bf project_{[0,1]}}(\mathbf{A}^{(i+1)})$  \;
    $i\leftarrow i+1$
 }
 \caption{Speaker diarization}
 \label{algo:diarization}
\end{algorithm}

The $\mathbf{shrink}$ operation on a matrix $\mathbf{X}$ is defined in equation~\ref{eq:shrinkage} where $\gamma$ and $\lambda$ represents the learning rate and the Lagrange multiplier respectively. This operation pushes each component of the matrix towards zero and thus achieves sparsity quickly~\cite{beck2009fast}.
\begin{equation}
    \mathbf{shrink}(\mathbf{X}):= sign(\mathbf{X})max(0, |\mathbf{X}| - \gamma\lambda)
    \label{eq:shrinkage}
\end{equation}
The $\mathbf{project_{unitdisk}}$ and $\mathbf{project_{[0,1]}}$ are two projection operations responsible for keeping the magnitudes of the model parameters in check. They are defined in equation~\ref{eq:project_unit_disk} and equation~\ref{eq:project_zero_one}, respectively.
\begin{equation}
    \mathbf{project_{unitdisk}}(\mathbf{X}):= \frac{\mathbf{X}[:,c]}{\|\mathbf{X}[:,c]\|_2} \quad\forall_{\text{column index}, c}
    \label{eq:project_unit_disk}
\end{equation}
\begin{equation}
    \mathbf{project_{[0,1]}}(\mathbf{X}):= \max{\{0, \min \left(1, \mathbf{X} \right) \}}
    \label{eq:project_zero_one}
\end{equation}
For $\lambda_1$, $\lambda_2$, and $\lambda_3$, we use the values $0.3366$, $0.2424$, and $0.06$ respectively. We obtained these values using a Bayesian Hyper-parameter search as implemented in the Google Cloud Platform~\cite{golovin2017google}. We used a randomly sampled subset (n=60) from the validation set.

\subsubsection{Determining the maximum number of speakers}
The use of $\ell_1$ norm of $\mathbf{\Psi}$ in equation~\ref{eq:optimization_objective} ensures that algorithm~\ref{algo:diarization} utilizes as few embedding vectors as possible to reconstruct the embedding signal, $\mathbf{\mathcal{E}}$. This process relieves the users from the burden of supplying an \emph{exact} number of speakers to the algorithm. However, setting a reasonable \emph{maximum} number of speakers ($k$) is still crucial because setting it too high would make the computation unnecessarily slow.
We obtain a reasonable value for $k$ from the estimated rank of the embedding signal. The columns of the embedding signal can be thought of just the embeddings of different speakers that are ``copied'' over time. In that scenario, it is possible to compute the number of the speakers by computing the \textit{Singular Value Decomposition} of the embedding signal and counting the number of non-zero singular values. However, the measurement noise in the speaker embeddings makes the situation a bit complicated because it yields many small but non-zero singular values. We circumvent this situation by sorting the singular values in a descending order and find the ``knee'' of the curve using the \emph{Kneedle}~\cite{satopaa2011finding} algorithm. We multiply the location of the knee by a factor of $2.5$ which gives us a useful margin of error for the upper bound for number of speakers and ensures that the sparsity constraint for the embedding basis matrix still holds.

\subsubsection{Overlap awareness}
\begin{figure}[t]
  \centering
  \includegraphics[width=0.9\linewidth]{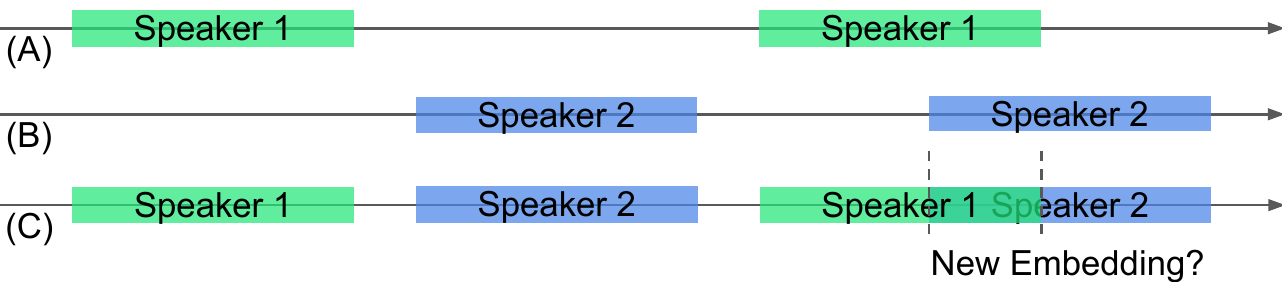}
  \caption{A practical situation where the proposed algorithm is able to recover the true speaking activity from an overlapping region}
  \label{fig:overlap_handling}
\end{figure}

The utilization of the sparsity constraint along with the linearity property as in equation~\ref{eq:vggvox_linearity} allows the proposed algorithm to practically identify and disambiguate regions with overlapping speakers. We describe the rationale for this capability with the help of figure~\ref{fig:overlap_handling}. Let us assume the rows \textit{A} and \textit{B} in figure~\ref{fig:overlap_handling} show the regions in the audio recording where speaker 1 and speaker 2 are speaking respectively. The resulting embedding signal is shown in row \textit{C}. Noticeably, there is a region in this row where the two speakers overlap. Therefore, as per equation~\ref{eq:vggvox_linearity}, the resulting embedding sequence in this overlapping region will be a linear combination of the embeddings from speaker 1 and speaker 2. 

There are two ways of accommodating this resulting embedding sequence using the embedding basis matrix and activity matrix. The overlapping region can be interpreted as a new speaker embedding with the corresponding activity value of $1.0$. Alternatively, the activity value for both speakers can be set to be nonzero without introducing a new speaker embedding. Since the introduction of a new embedding would incur additional loss, the optimization algorithm will prefer the second approach. There are settings in which this algorithm will fail to identify overlapping regions (e.g. when a speaker always is overlapped, or when the resulting linear combination of speaker embeddings happens to exactly match the embedding of another speaker), but these are unlikely to occur in practice.

\section{Experiments}

\begin{table*}[th]
  \caption{Diarization results on two data sets}
  \label{tab:resultsboth}
  \centering
  \begin{tabular}{ c|cccc|cccc }
    \toprule
   &  \multicolumn{4}{|c}{\textit{VoxConverse} data set} &   \multicolumn{4}{|c}{\textit{This American Life} Podcast} \\
    \midrule
\multicolumn{1}{c}{\textbf{Method}} &   \multicolumn{1}{|c}{\textbf{DER}}    &   \multicolumn{1}{c}{\textbf{Purity}} &   \multicolumn{1}{c}{\textbf{Coverage}}   &   \multicolumn{1}{c}{\textbf{F}} &   \multicolumn{1}{|c}{\textbf{DER}}    &   \multicolumn{1}{c}{\textbf{Purity}} &   \multicolumn{1}{c}{\textbf{Coverage}}   &   \multicolumn{1}{c}{\textbf{F}} \\
    \midrule
    GCP & $0.39$ & $0.74$ & $0.92$ & $0.80$~~~     & $0.63$ & $0.47$ & $0.93$ & $0.62$~~~ \\
    Spectral Clust.& $0.50$ & $0.70$ & $0.87$ & $0.76$~~~ & $0.53$ & $0.73$ & $0.77$ & $0.73$~~~ \\
    PyAnnote v1.1 & $0.42$ & $0.68$ & $0.92$ & $0.76$~~~ & $0.38$ & $0.73$ & $0.75$ & $0.74$~~~ \\
    PyAnnote v2.0 & $0.12$ & $0.90$ & $0.92$ & $0.91$~~~ & $0.25$ & $0.94$ & $0.92$ & $0.93$~~~ \\
    \midrule
    One Speaker & $0.53$ & $0.57$ & $1.00$ & $0.70$~~~  & $0.77$ & $0.32$ & $1.00$ & $0.47$~~~ \\
    Random & $0.67$ & $0.57$ & $0.52$ & $0.52$~~~ & $0.83$ & $0.32$ & $0.48$ & $0.37$~~~ \\
    \midrule
    \textbf{Sparse Opt. (Ours)} & $0.56$ & $0.90$ & $0.60$ & $0.70$~~~ & $0.35$ & $0.84$ & $0.84$ & $0.83$~~~ \\ 
    \bottomrule
  \end{tabular}
\end{table*}

In order to ground the diarization performance of the proposed algorithm, we evaluate it against a few publicly available speaker diarization solutions. We conduct the evaluation over the two largest publicly available annotated datasets: \emph{This American Life Podcast Dataset}~\cite{mao2020speech} and \emph{VoxConverse}~\cite{chung2020spot}.

The ``This American Life Podcast'' dataset~\cite{mao2020speech} consists of  $663$ episodes from a podcast with the same name. The dataset consists of dialog transcripts and human annotated speaker labels at the utterance level. It consists of a total of $637$ hours of audio. On average, each audio recording is $57.7$ minutes long and contains 18 speakers in this dataset. 

The ``VoxConverse'' dataset~\cite{chung2020spot} consists of 448 audio recordings from various speakers. This dataset was created by mixing together a collection of audio samples from YouTube videos. The speaker labels were obtained using an automated pipeline that can recognize the speakers' face, and can associate the audio with the speaker by analyzing the facial movements~\cite{chung2020spot}. The average audio length in VoxConverse is $8.5$ minutes and contains $5.5$ speakers per audio.

We compare the proposed diarization algorithm over several other approaches as shown in Table~\ref{tab:resultsboth}. \textbf{GCP} is a commercial baseline representing the speaker diarization service provided by Google Cloud Platform 
as of March 8\textsuperscript{th}, 2022. \textbf{Pyannote} 
is a Python open-source toolkit which also provides trained diarization models~\cite{9052974}. 
PyAnnote is a neural-network-based supervised approach for diarization currently released as version 1.1. 
A successor, PyAnnote v2.0, is currently actively under development with the technical details undisclosed at the time of writing. 
\textbf{Spectral Clustering} is an in-house implementation of the classical diarization pipeline using the VggVox~\cite{nagrani2020voxceleb} embedding and spectral clustering. Comparison against this baseline provides an indication of how much of the change in the evaluation metrics is caused by the sparse optimization algorithm alone, and not by the improved discriminative capability of the embeddings. We also define two naive baselines to evaluate the lowest bar for diarization:  \textbf{One Speaker}, where every word in the transcript is assigned to a single speaker, and \textbf{Random}, where every word in the transcript is randomly assigned to some speaker.

For evaluation, we use some standard diarization performance metrics from the Pyannote metrics~\cite{9052974} python library. \textbf{Diarization Error Rate (DER)} is the sum of the duration of non-speech regions incorrectly classified as speech (\emph{false alarm}), the duration of speech regions incorrectly classified as non-speech (\emph{missed detection}), and the duration of speaker \textit{confusion}, as a ratio of the total duration of speech for all speakers.

\textbf{Purity} is a precision-related measure representing the quality of the each \textit{predicted} speech segments. It is represented by equation~\ref{eq:purity}, where  $|\text{cluster}|$ and $|\text{speaker}|$ represent the speech duration of the predicted speech segments and the reference speech segments respectively. $|\text{cluster } \cap \text{ speaker}|$ represents the duration of their intersection. \textbf{Coverage} is the corresponding recall-related measure calculated for each \textit{reference} speech segment as in equation~\ref{eq:coverage}. The \textbf{F} score is the harmonic mean of purity and coverage. A better system has lower DER, or higher purity, coverage, or F score.
\begin{equation}
    \text{purity} = \frac{\sum_{\text{cluster}}\max_{\text{speaker}}|\text{cluster } \cap \text{ speaker}|}{\sum_{\text{cluster}}|\text{cluster}|}
    \label{eq:purity}
\end{equation}
\begin{equation}
    \text{coverage} = \frac{\sum_{\text{speaker}}\max_{\text{cluster}}|\text{ speaker } \cap \text{ cluster}|}{\sum_{\text{speaker}}|\text{speaker}|}
    \label{eq:coverage}
\end{equation}

It is evident from Table~\ref{tab:resultsboth} that the sparse optimization algorithm works much better in the This American Life podcast dataset than the VoxConverse dataset. This behavior is due to the fact that VoxConverse audio recordings are much shorter and, therefore, result in fewer columns in the embedding signal as compared to the rows. This transforms the matrix factorization problem from the domain of \emph{under-determined} systems to an \emph{over-determined} system that violates the assumptions for sparse optimization formulation. However, we are interested in diarization solutions for the podcast applications---which are typically long, and therefore suitable for our use case.

The overall diarization purity for the proposed algorithm is better than GCP, spectral clustering, and PyAnnote v1.1 by a large margin: the predicted clusters are more pure, i.e. not a mix of speakers, at a certain cost in coverage. The overall performance (F-score) also shows similar trend. It is second only to PyAnnote v2.0---for which the algorithm is still undisclosed. If PyAnnote v2.0 works by taking a supervised approach it has a risk of being susceptible to domain mismatch. In addition, supervised approaches are often difficult to maintain due to difficulties in collecting data and the amount of engineering effort it takes to keep the dataset balanced for preventing bias and ensuring fairness.

\section{Conclusions}
We present a novel method for speaker diarization that is unsupervised---does not require human annotations which are difficult to collect and maintain. The algorithm performance beats a commercial solution (GCP) across all standard metrics. It is completely audio-based, therefore, agnostic to transcription or any other language-dependent processing. In addition, it removes the burden of supplying an exact number of speakers from the users; and thus works in a tune-free or self-tuned fashion. Accommodating for the overlapping speech is built into the design of the estimator, therefore removes the necessity for any ad-hoc post-processing. Being a first order optimization approach that is implemented in TensorFlow, this algorithm is massively parallelizable and has great potential for speed improvement on the right kind of processing infrastructure (e.g. utilizing TPU's or GPU's). 

\bibliographystyle{IEEEtran}
\bibliography{sparse_diarization}

\balance
\end{document}